\documentclass{article} 
\usepackage{iclr2024_conference,times}

\usepackage{amsmath,amsfonts,bm}

\def\eqref#1{equation~\ref{#1}}

\def\1{\bm{1}}

\DeclareMathAlphabet{\mathsfit}{\encodingdefault}{\sfdefault}{m}{sl}
\SetMathAlphabet{\mathsfit}{bold}{\encodingdefault}{\sfdefault}{bx}{n}

\usepackage{hyperref}
\usepackage{cleveref}
\Crefname{equation}{Equation}{Equations}
\usepackage{url}
\usepackage{algorithm}
\usepackage{algorithmic}
\usepackage{subfigure}
\usepackage{graphicx}
\usepackage{diagbox}
\usepackage{tabularx}
\usepackage{svg}
\usepackage{booktabs} 
\usepackage{siunitx} 
\usepackage{multirow}
\usepackage{hyphenat}
\usepackage{wrapfig}
\usepackage[normalem]{ulem}

\title{FL-TAC: Enhanced Fine-Tuning in Federated Learning via Low-Rank, Task-Specific Adapter Clustering}

\author{Siqi Ping$^1$\thanks{Siqi Ping and Yuzhu Mao contribute equally. Wenbo Ding is the corresponding author.}, Yuzhu Mao$^1$\footnotemark[1], Yang Liu$^{2,3}$, Xiao-Ping Zhang$^{1,5}$ \& Wenbo Ding$^{1,3,4}$\footnotemark[1] \\
$^1$Tsinghua-Berkeley Shenzhen Institute, \\ \hspace{2pt} Tsinghua Shenzhen International Graduate School, Tsinghua University \\
$^2$ Institute for AI Industry Research (AIR), Tsinghua University\\
$^3$ Shanghai AI Lab, Shanghai, China\\
$^4$ RISC-V International Open Source Laboratory, Shenzhen, China\\
$^5$ Department of Electrical, Computer and Biomedical Engineering, Ryerson University \\
\texttt{psq22@mails.tsinghua.edu.cn, myz20@tsinghua.org.cn} \\ 
\texttt{liuy03@air.tsinghua.edu.cn, xzhang@ee.ryerson.ca} \\
\texttt{ding.wenbo@sz.tsinghua.edu.cn} \\
}

\newcolumntype{M}[1]{>{\centering\arraybackslash}m{#1}}

\iclrfinalcopy 
\begin{document}

\maketitle

\begin{abstract}

Although large-scale pre-trained models hold great potential for adapting to downstream tasks through fine-tuning, the performance of such fine-tuned models is often limited by the difficulty of collecting sufficient high-quality, task-specific data. Federated Learning (FL) offers a promising solution by enabling fine-tuning across large-scale clients with a variety of task data, but it is bottlenecked by significant communication overhead due to the pre-trained models' extensive size. This paper addresses the high communication cost for fine-tuning large pre-trained models within FL frameworks through \textit{low-rank fine-tuning}. Specifically, we train a low-rank adapter for each individual task on the client side, followed by server-side clustering for similar group of adapters  to achieve task-specific aggregation. Extensive experiments on various language and vision tasks, such as GLUE and CIFAR-10/100, reveal the evolution of task-specific adapters throughout the FL training process and verify the effectiveness of the proposed low-rank \underline{\textbf{t}}ask-specific \underline{\textbf{a}}dapter \underline{\textbf{c}}lustering (TAC) method. 



\end{abstract}

\section{Introduction}

Large-scale pre-trained models, such as Large Language Models (LLMs) trained on extensive data, demonstrate superior performance in natural language processing and remarkable adaptability to various downstream tasks~\citep{brown2020language, ouyang2022training, touvron2023llama,zhang2022opt,dosovitskiy2020image,brohan2023can}. However, fine-tuning these models for specific tasks necessitates high-quality, task-relevant data, posing challenges for edge users with diverse, yet limited, private datasets~\citep{zhang2023towards}. In this context, Federated Learning (FL) offers a solution by enabling collaborative model training across devices with distributed data, achieving comparable results to centralized training with abundant data~\citep{mcmahan2017communication}.

Despite its advantages, FL faces significant communication overhead due to frequent exchanges of models or gradients between distributed clients and central servers~\citep{lim2020federated,zhao2023towards}. This issue becomes particularly critical when fine-tuning LLMs that have a vast number of parameters~\citep{yi2023fedlora}. Consequently, Parameter-Efficient Fine-Tuning (PEFT) methods, such as Low-Rank Adaptation (LoRA), are essential in FL environments to reduce communication costs~\citep{houlsby2019parameter,lester2021power,li2021prefix,zaken2022bitfit,fu2023effectiveness,hu2021lora}. For example, LoRA leverages the inherent low-dimensional structure of over-parameterized models by training only a low-rank adapter with fewer parameters during fine-tuning~\citep{hu2021lora}. Nevertheless, existing research has indicated a trade-off: A lower number of trainable parameters can lead to increased approximation errors and reduced model performance for downstream tasks, particularly when the training target is complex, for instance, when there is a significant optimization error between the target model and the model before fine-tuning~\citep{zeng2023expressive}.


Addressing the challenges of low-rank fine-tuning in FL environments catering to diverse tasks, this paper investigates:

\begin{center}
\textit{How to save communication costs through low-rank fine-tuning \\
while still enabling effective task adaptation for each client?}
\end{center}

A natural solution to this problem is to customize the low-rank adapter for each task on the client side and aggregate the adapters for the same task on the server side. To this aim, this paper examines the evolution and interrelationships of low-rank adapters from distributed clients for different tasks. The findings are employed to enhance FL aggregation for multi-task objectives. Specifically, the proposed FL process with low-rank \underline{\textbf{t}}ask-specific \underline{\textbf{a}}dapter \underline{\textbf{c}}lustering (FL-TAC) allows each client to train and upload a unique, extremely low-rank adapter for each task. The server subsequently performs a clustering step to differentiate adapters for different tasks and aggregates the adapters with the same task target.

The contributions of this paper are threefold: 1) The proposed FL-TAC algorithm enables an efficient and effective FL framework, which is capable of accommodating various target tasks from image to text classification and generation, and outperforms single-adapter baselines in both performance and communication efficiency. 2) The successful clustering of task-specific adapters provides insights into the evolution of trainable parameters during the FL training process, demonstrating how this property can be utilized for effective generalization across multiple downstream tasks in a distributed setting. 3) Extensive experiments on text datasets such as the GLUE benchmark and image datasets like CIFAR-10/100 demonstrate high clustering accuracy of FL-TAC algorithm and its general applicability to a broad range of downstream tasks within a unified framework.

The rest of this paper is organized as follows: Section 2 briefly outlines previous work related to this paper. In Section 3, the proposed algorithm is described in detail. The experimental results are provided in Section 4, and the concluding remarks are summarized in Section 5.

\begin{figure}[h]
\begin{center}
\subfigure[Server-Client adapter exchange.]{
\includegraphics[width=0.4\textwidth]{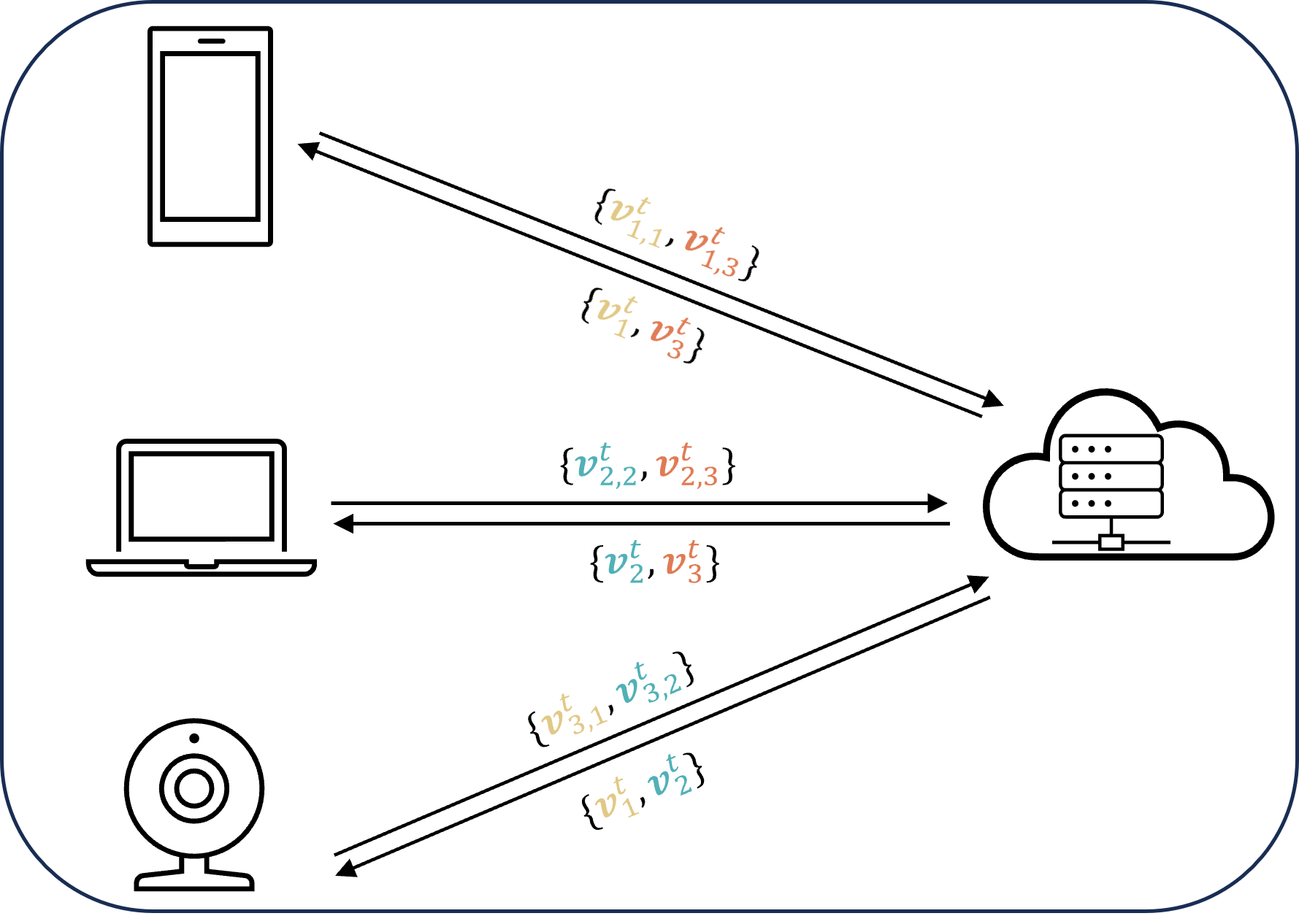}
\label{overview}
}\subfigure[Server clustering and aggregation.]{
\includegraphics[width=0.4\textwidth]{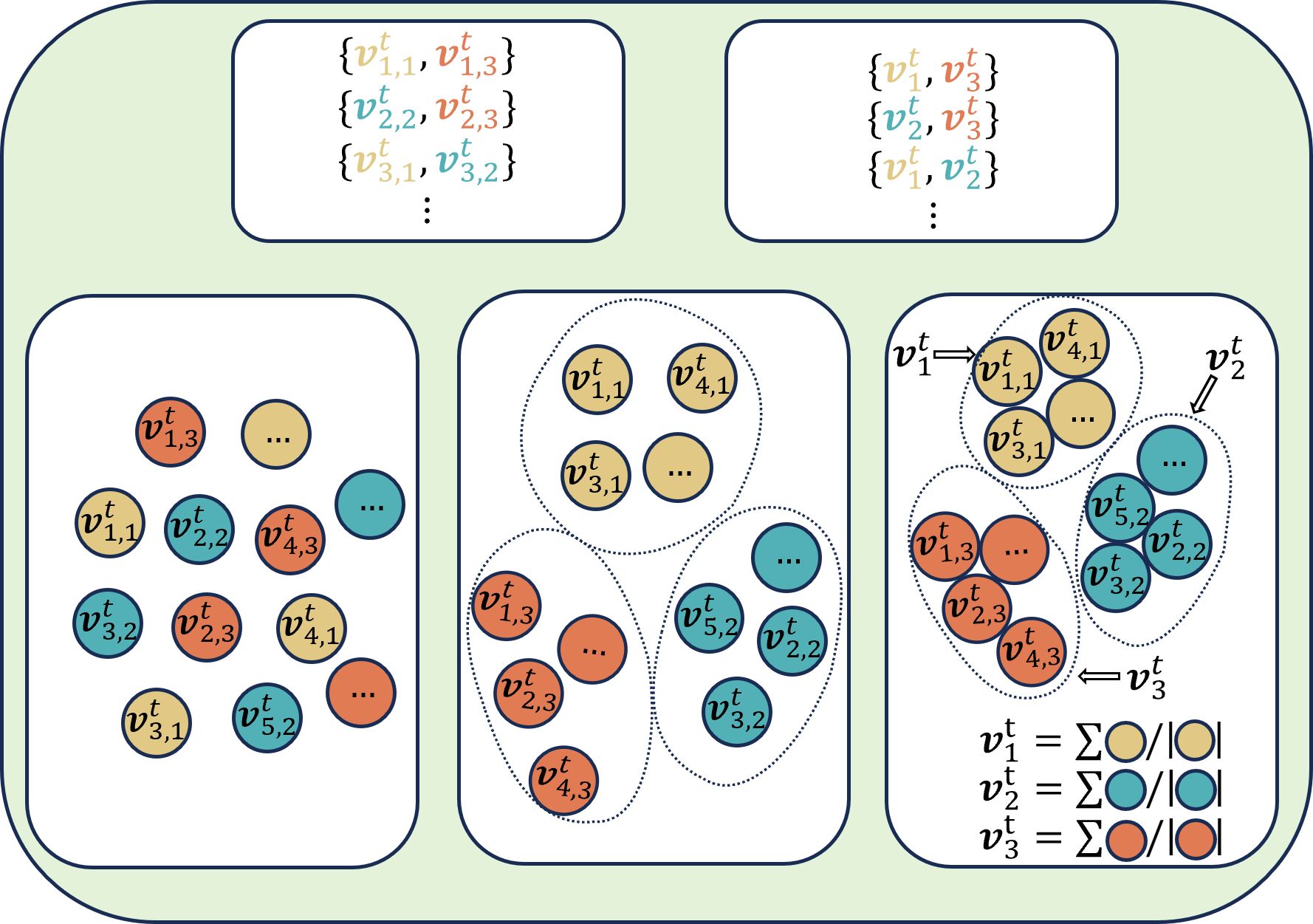}
\label{aggregation}
}
\end{center}
\caption{FL-TAC framework.}
\label{aggregation_complete}
\end{figure}

\vspace{-10pt}

\section{Related Work}

\subsection{FL for Fine-tuning Large-scale Pre-trained Models}

The fine-tuning of large-scale pre-trained models in FL scenarios has been recently studied, leading to several significant contributions particularly relevant to this work. A line of research, represented by Federated Instruction Tuning (FedIT), has made substantial progress in fine-tuning LLMs in FL scenarios~\citep{zhang2023towards}. FedIT offers a generalizable experimental framework that utilizes distributed instruction data sources to enhance the final fine-tuned model's performance. This work has laid essential groundwork in the field, demonstrating the potential of FL in the context of fine-tuning LLMs. Following this, the FedLoRA approach has enabled efficient FL fine-tuning and communication through the PEFT methods~\citep{yi2023fedlora}. PEFT methods finetune pre-trained models with reduced trainable parameters, and can be broadly categorized into addition-based methods, specification-based methods, and reparameterization-based methods as follows~\citep{zhou2023opportunities}:

\textbf{Addition-based Method.} The core idea of addition-based methods is to integrate extra trainable modules into the pre-trained model. Two representatives of addition-based methods are adapter-based tuning~\citep{houlsby2019parameter} and prompt-based tuning~\citep{lester2021power,li2021prefix}. The former introduces small, task-specific neural networks as adapters between the layers of the pre-trained model, while the latter incorporates additional context by wrapping the original input. In this way, the number of trainable parameters during fine-tuning is limited to the size of the added adapters or input context.

\textbf{Specification-based Method.} This type of PEFT methods freezes most of the pre-trained model parameters and updates only a selected subset of parameters during fine-tuning. For instance, BitFit~\citep{zaken2022bitfit}, a sparse finetuning method, only adjusts the bias terms in neural networks, and the Second-order Approximation Method (SAM) identifies the subset of parameters to be adjusted by solving an optimization problem ~\citep{fu2023effectiveness}.

\textbf{Reparameterization-based Method.} This significant line of research is inspired by previous observations that the performance yielded by a low-dimensional matrix can match that of a fine-tuned deep layer in large-scale pre-trained models~\citep{aghajanyan2021intrinsic, karimi2021compacter}. In this context, LoRA is a widely-used reparameterization-based algorithm that takes advantage of the existence of a low intrinsic rank to reduce trainable parameters~\citep{hu2021lora}.

In this paper, we adopt the well-recognized LoRA method, similar to FedLoRA, for its wide empirical success and robust theoretical guarantee~\citep{zeng2023expressive}. It should be noted that the current analysis and experimental results can be easily extended to other PEFT methods by substituting the LoRA adapters with the training outcomes derived from other PEFT techniques.

While previous works mark significant advances in the field of efficient fine-tuning through FL, the challenges arising from task diversity remain underexplored and inadequately addressed. Considering the motivation of FL, which is to utilize distributed data sources, it is natural for each unique data source, such as an edge device or user, to have diverse local tasks to handle. Consequently, there remains a need for an FL framework that enables parameter-efficient training and aggregation among clients with different optimization targets to approximate.

\subsection{FL for Multi-Task Learning}
Multi-task learning involves the simultaneous training on multiple related tasks to leverage their commonalities or differences for improved performance over all tasks. Methods to achieve multi-task learning in FL scenarios include federated transfer learning~\citep{liu2020secure,gao2019privacy} and clustered FL~\citep{ghosh2020efficient, sattler2020clustered, briggs2020federated}. Among these approaches, federated transfer learning aims to transfer information learned from one task to another, while clustered FL focuses on classifying distributed clients into multiple clusters based on certain task representations prior to the aggregation step. This pre-aggregation clustering improves multi-task performance by enhancing the information sharing within clients handling the same task. Existing studies in clustered FL have explored various forms of task representation, including models, gradients, logits, and latent representations~\citep{kulkarni2020survey,mansour2020three,mao2023safari,zhu2021data}.

Given its ability to classify clients according to the underlying tasks, clustered FL aligns more closely with the objective of this study, which is to customize multiple task-specific low-rank local adapters for each client within the FL framework and to perform aggregation within the same task. Therefore, clustered FL is employed in this paper to enhance performance for each downstream task involved in an FL training process. Moreover, the recent emergence of PEFT methods driven by LLMs fine-tuning has yet to be incorporated in previous clustered FL studies, leaving a gap in the current literature for a clustered FL framework for multi-task learning that accommodates task representations provided by PEFT methods.

Based on the above analysis, this study aims to address, to some extent, the limitations and gaps identified in the aforementioned works. The algorithm introduced in Section 3 offers a novel approach to managing multiple tasks per client during the efficient fine-tuning of large pre-trained models within FL frameworks.


\vspace{-5pt}

\section{Method}

\subsection{Overview of Fine-tuning through FL}

Consider an FL scenario involving a central server and $m$ clients, collectively addressing a total of $N$ downstream tasks. Each client $i$ in the client set $\mathcal{M}=\{1,2,\ldots,m\}$ is responsible for a subset of these tasks, with a total of $n_i \leq N$ local tasks to handle. Thus, the local dataset for client $i$ can be denoted as $\mathcal{D}_i=\{ \mathcal{D}_i^j \vert j \in \mathcal{N}_i\}$, where each $\mathcal{D}_i^j$ represents a downstream task dataset of client $i$, and $\mathcal{N}_i$ denotes the indices of  client $i$'s local tasks within the global set of tasks.

In centralized learning scenarios, each client $i$ fine-tunes a pre-trained model using its local dataset $\mathcal{D}_i$. However, FL leverages the collective dataset $\mathcal{D}_{\mathcal{M}} = \bigcup_{i=1}^m \mathcal{D}_i$ as a comprehensive, global dataset to provide richer information. This is achieved by optimizing the global function defined as:
\begin{align}
    \min_{\bm{\theta}} {\mathcal{F}}\left(\bm{\theta}\right) = \sum_{i=1}^{m} \frac{|\mathcal{D}_i|}{|\mathcal{D}_{\mathcal{M}}|} \mathcal{F}_i\left(\bm{\theta}\right), \nonumber \\
     \mathcal{F}_i(\bm{\theta}) := \mathbb{E}_{(\bm{x},y) \sim \mathcal{D}_i} \left[\mathcal{L} (\bm{\theta}; (\bm{x}, y))\right].
     \label{Equation1}
\end{align}
Here, $(\bm{x},y)\sim \mathcal{D}_i$ denotes data-label pairs sampled from client $i$'s local dataset $\mathcal{D}_i$. $|\mathcal{D}_i|$ and $|\mathcal{D}_{\mathcal{M}}| := \sum_{i\in \mathcal{M}} |\mathcal{D}_i|$ represent the sizes of the local and global datasets, respectively. The function $\mathcal{F}_i(\cdot)$ represents the expected local objective concerning client $i$'s dataset, a specific loss function $\mathcal{L}(\cdot)$, and the global model parameters $\bm{\theta}$. The global objective $\mathcal{F}(\cdot)$ is a weighted average of these local objectives, with weights proportional to the size of each client’s dataset. The optimal global model $\bm{\theta}^*$ that minimizes this function is expected to perform well on the global dataset $\mathcal{D}$.

In practice, during the $t$-th communication round, the server first selects a subset of clients $\mathcal{M}_t$ with total number of $m_t$ clients, either randomly or based on specific criteria, and broadcasts the current global model, denoted as $\bm{\theta}^{t-1}$, to these clients. Each chosen client $i$ in $\mathcal{M}_t$ then updates this model $\bm{\theta}^{t-1}$ through $\tau$ local training steps to obtain its local model $\bm{\theta}_{i}^{t}$. This is typically done using the update rule $\bm{\theta}_{i, k}^{t} = \bm{\theta}_{i, k-1}^{t} - \eta \nabla{\bm{\theta}_{i, k-1}^{t}} \mathcal{F}_i(\bm{\theta}_{i, k-1}^{t})$ for $k=1,\ldots,\tau$, where $\eta$ is the learning rate and $\bm{\theta}_{i, 0}^{t} = \bm{\theta}^{t-1}$. Post local training, clients send their updated local models $\bm{\theta}_i^{t} = \bm{\theta}_{i, \tau}^{t}$ back to the server, which then aggregates them into the updated global model $\bm{\theta}^{t} = \sum_{i \in \mathcal{M}_t} \frac{|\mathcal{D}_i|}{|\mathcal{D}_{\mathcal{M}_t}|} \bm{\theta}_i^{t}$. This cycle repeats until $T$ rounds are completed, resulting in the final model $\bm{\theta}^{T}$ that approximates the optimal~$\bm{\theta}^*$.

However, for initial global models like LLMs that have a significant scale, fine-tuning through the FL process becomes infeasible due to the heavy communication overhead from transmitting large-scale models. As a solution, the LoRA technique can be incorporated to reduce the number of transmitted parameters~\citep{hu2021lora}. Specifically, for an original weight matrix $\bm{W} \in \mathbb{R}^{d\times k}$ in a pre-trained model and its corresponding target weight matrix $\bm{W}' \in \mathbb{R}^{d\times k}$ after fine-tuning, LoRA approximates the fine-tuned output $\bm{W}'\bm{z}$ for an input $\bm{z}$ using a low-rank decomposition: $\bm{W}'\bm{z} = \bm{W}\bm{z} + \bm{B}\bm{A} \bm{z}$, where $\bm{B}\in\mathbb{R}^{d\times r}$, $\bm{A}\in \mathbb{R}^{r\times k}$, and the rank $r \ll \min(d;k)$. This allows the original weight matrix $\bm{W}$ to remain unchanged during fine-tuning, with only the smaller matrices~(also termed as adapters) $\bm{A}$ and $\bm{B}$ being adjustable.

By incorporating the LoRA technique, the large-scale pre-trained model, denoted as $\bm{\theta}$, only needs to be transmitted from the central server to the clients at the beginning of the first communication round. In subsequent rounds (i.e., the $t$-th round), the server first sends a global adapter $\bm{v}^{t-1}$, and each client $i$ updates $\bm{v}^{t-1}$ through efficient local fine-tuning to obtain the local adapter $\bm{v}_i^{t} = [\bm{A}_i^0, \bm{B}_i^0, \ldots, \bm{A}_i^L, \bm{B}_i^L]$ for a total of $L$ weight matrices in the pre-trained model $\bm{\theta} = [\bm{W}^0, \ldots, \bm{W}^L]$. Global aggregation is subsequently performed using the local adapters provided by the clients, resulting in $\bm{v}^{t} = \sum_{i \in \mathcal{M}_t} \frac{|\mathcal{D}_i|}{|\mathcal{D}_{\mathcal{M}_t}|} \bm{v}_i^{t}$, and the new global model can be approximated as $\bm{\theta}^{t} = \bm{\theta}^{t-1} + \bm{v}^{t}$.

\subsection{Core Concepts and Approach}

\vspace{-10pt}


\begin{figure}[t]
\begin{center}
\subfigure[Proportion of task data among clients.]{
\includegraphics[width=0.5\textwidth]{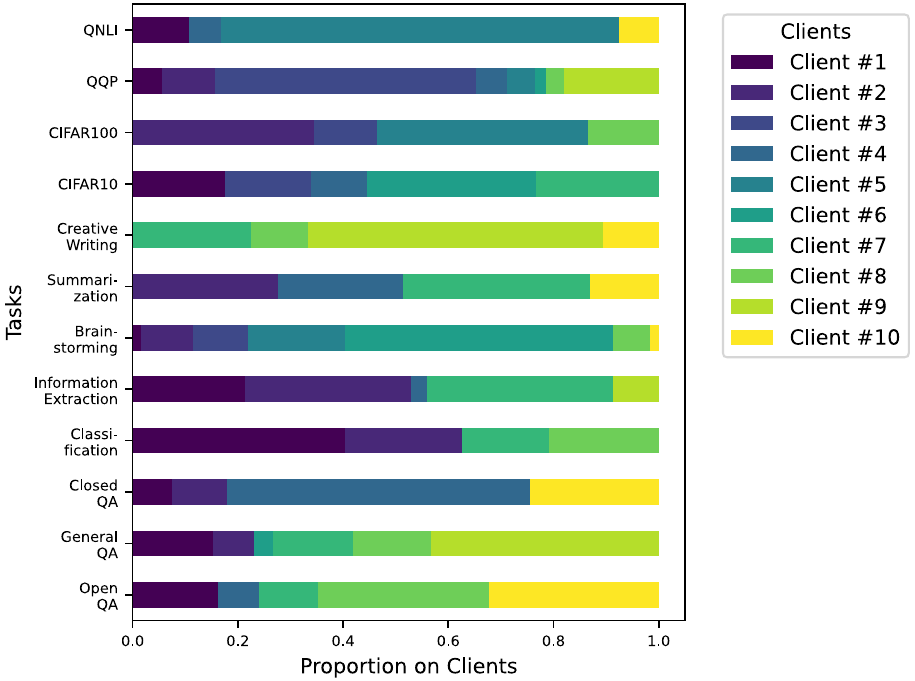}
\label{data distribution}
}\subfigure[Scores for each task in Databricks-Dolly-15k.]{
\includegraphics[width=0.5\textwidth,height=5cm]{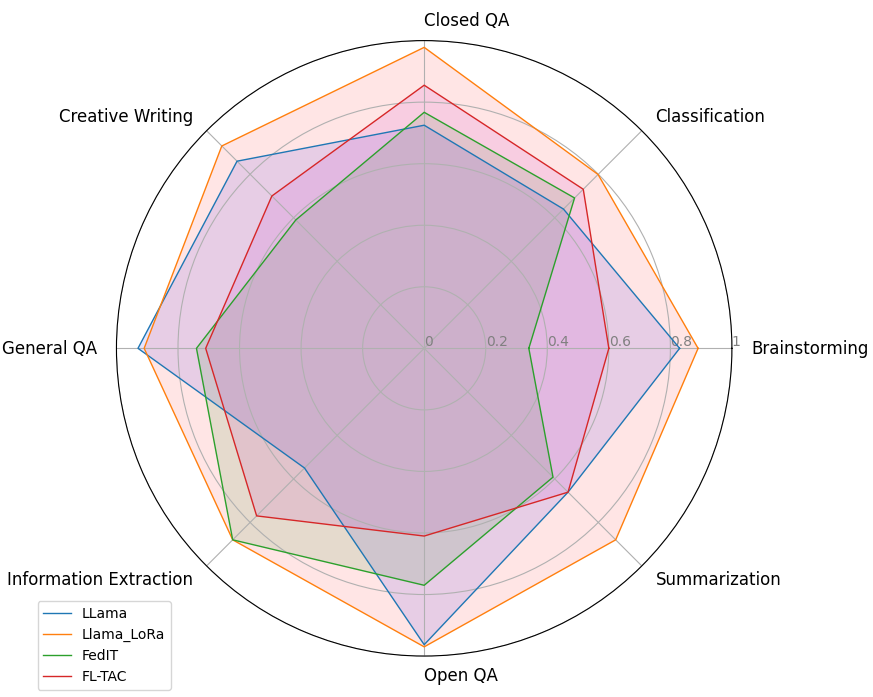}
\label{radar}
}
\end{center}
\caption{Visualization of data distribution across clients and performance evaluation on Databricks-Dolly-15k multitask dataset.}
\end{figure}

In the general process of fine-tuning large pre-trained models through FL, each client $i$ typically leverages a single LoRA adapter for downstream task adaptation, despite having multiple local tasks. Achieving empirical success is feasible due to the exceptional emergent abilities provided by recent pre-trained models like GPT and Llama~\citep{brown2020language, touvron2023llama}. However, recent theoretical research indicates that the expressive power of fine-tuned models can be significantly restricted by the aggressive reduction of LoRA rank, such as 16 or even 8, as often seen in literature~\citep{zeng2023expressive}. This presents an additional challenge to fine-tuning via FL, where such aggressive use of the LoRA method is considered necessary.

In the context of fine-tuning large models through FL, the key of this work lies in training a low-rank task-specific adapter for each individual task at the client side. The motivation arises from the observation that when the LoRA rank falls below a certain threshold, the approximation error begins to significantly increase.  
For tasks with higher complexity (measured by the approximation error before fine-tuning), reducing the LoRA rank below this threshold results in a more rapid increase in the approximation error, as shown in previous study~\citep{zeng2023expressive}. 

Intuitively, using a single adapter for multiple tasks results in a higher approximation error before fine-tuning and thus a more rapid and significant increase in the approximation error caused by LoRA rank reduction. Conversely, using a unique adapter for each individual task leads to a slower \begin{wrapfigure}[15]{r}{0.5\textwidth}
  \centering
  \vspace{-12pt}
  \includegraphics[width=0.5\textwidth]{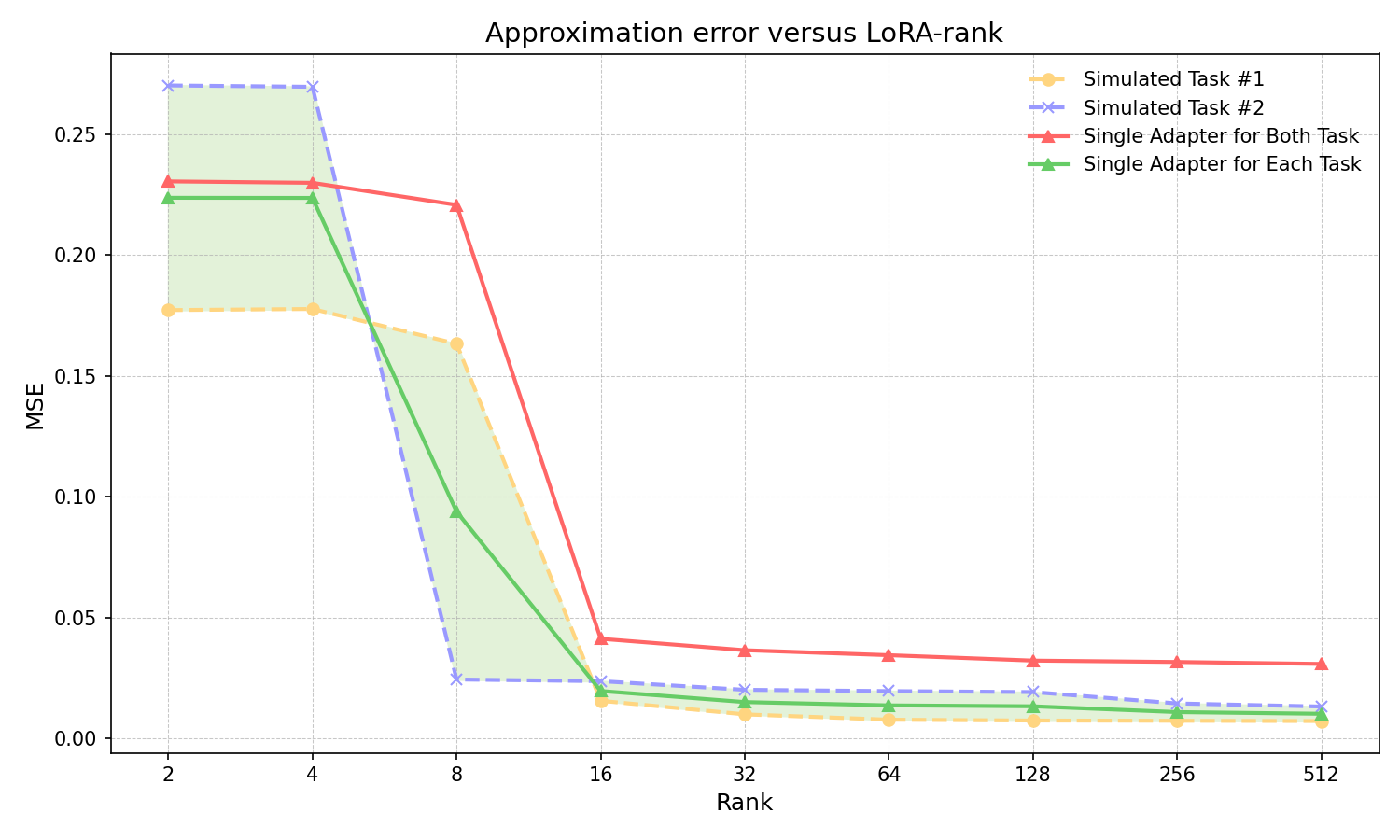}
  \vspace{-15pt}
  \caption{Approximation error (measured by MSE) versus LoRA-rank.}
  \label{fig:simulation}
\end{wrapfigure} and more gradual increase in the approximation error. To validate this intuition, a toy simulation is conducted in this study. In this simulation, data points in simulated task 1 are randomly initialized scalars with labels created by applying a noisy sinusoidal transformation to the data points and adding Gaussian noise. Data points in simulated task 2 are generated similarly, except for using a noisy sinusoidal transformation with a phase shift of 0.5 and Gaussian noise of higher variance. 
A Fully Connected Neural Network (FNN) with an adaptable hidden layer size is employed to simulate the LoRA adapter, train the model, and compute the generalization error (measured by Mean Squared Error (MSE), also referred to as approximation error to maintain consistency with previous work) after 1000 epochs. Figure~\ref{fig:simulation} shows preliminary results indicating that training an FNN for each simulated task consistently yields a lower average approximation error compared to training an FNN for both simulated tasks, particularly when the LoRA rank (simulated by the hidden layer size of the FNN) is low (e.g., below 16). This observation is particularly relevant when considering the fine-tuning/communication costs closely associated with the LoRA rank. For instance, when the LoRA rank $r$ is below 8, training two adapters with rank lower than $r/2$ yields a lower average approximation error compared to training one adapter with rank $r$ for both tasks.
      
Based on this finding, we propose training task-specific adapters for each client, aiming for an overall strategy that reduces the total number of trainable parameters while maintaining or reducing the average approximation error across all tasks. With this revised approach to fine-tuning via FL, each client $i$ must transmit multiple adapters to the central server in each communication round. To facilitate aggregation over adapters without explicit task labels from clients, a \emph{clustering step} is introduced on the server side to group adapters corresponding to the same task. This clustering step is essential to facilitate knowledge exchange and aggregation among clients handling the same task in the proposed FL fine-tuning process, thereby enhancing overall performance for that task.

\subsection{The FL-TAC Algorithm}

Building upon the introduced concepts and approaches, the proposed FL-TAC method enables each client \(i\) to fine-tune \(n_i\) distinct local adapters for its \(n_i\) local tasks. The final goal is to acquire \(N\) distinct global adapters for the total \(N\) downstream tasks in the system through FL iterations between local fine-tuning and global aggregation. In this way, the global objective is rewritten as:
\begin{align}
    \min_{\{\bm{v}_1,\ldots,\bm{v}_N\}}{\mathcal{\hat{F}}(\bm{v}_1,\ldots,\bm{v}_N)} &= \sum_{i=1}^{m} \frac{|\mathcal{D}_i|}{|\mathcal{D}_{\mathcal{M}}|} \mathcal{\hat{F}}_i\left(\bm{v}_1,\ldots,\bm{v}_N\right), \nonumber \\
     \mathcal{\hat{F}}_i\left(\bm{v}_1,\ldots,\bm{v}_N\right) &:= \sum_{j \in \mathcal{N}_i} \mathcal{\hat{F}}_{i, j}(\bm{v}_{j}, D_i^j) \nonumber \\
     &:= \sum_{j \in \mathcal{N}_i} \mathbb{E}_{(\bm{x},y) \sim D_i^j} \left[\mathcal{\hat{L}}(\bm{\theta}; \bm{v}_j; (\bm{x}, y))\right],
     \label{Equation2}
\end{align}
where \(\mathcal{\hat{F}}_i(\cdot)\) denotes client \(i\)'s local objective, constructed by the task-specific objective \(\mathcal{\hat{F}}_{i, j}(\cdot)\) for each task \(j \in \mathcal{N}_i\), and \(\mathcal{\hat{L}}\) refers to specific loss functions. It's important to note that \(\mathcal{\hat{F}}_{i, j}(\cdot)\) is influenced only by the task-specific trainable adapter \(\bm{v}_{j}\) and the underlying task data distribution \(D_i^j\), as the pre-trained model \(\bm{\theta}\) remains frozen during fine-tuning.

\begin{algorithm}
    \caption{FL-TAC}
    \label{alg:algorithm}
    \textbf{Inputs}: Pre-trained model $\bm{\theta}$; Client set $\mathcal{M}$; Local datasets $D_{i}^{j}$ for $i \in \mathcal{M}, j \in \mathcal{N}_i$. \\
    \textbf{Parameters}: Number of communication rounds $T$; Number of local steps $\tau$; Number of downstream tasks $N$.
    \begin{algorithmic}[1] 
    \STATE Server randomly initializes a global adapter $\bm{v}^{0}$.
         \STATE Server broadcasts the initial global adapter $\bm{v}^{0}$ to $\mathcal{M}$.
         \FOR{client $i$ in $\mathcal{M}$ in parallel}
        \STATE Initializes local task-specific adapters $\bm{v}_{i, j}^{0}=\bm{v}^{0}$ for each local task $j \in \mathcal{N}_i$.
        \ENDFOR
        \FOR{$t=1$ to $T$}
        \STATE Server selects subset of clients $\mathcal{M}_t$ for round $t$.
        \FOR{client $i$ in $\mathcal{M}_t$ in parallel}
        \STATE Performs \textit{FL-TAC Local Fine-tunin}g($\bm{\theta}, \tau, \{\bm{v}_{i,j}^{t-1} \vert j\in \mathcal{N}_i\}$).
        \STATE Sends updated local task-specific adapters $\bm{v}_{i,j}^{t}$for each local task $j \in \mathcal{N}_i$ to the server.
        \ENDFOR
        \STATE Server performs \textit{FL-TAC Global Clustering and Aggregation}($\{\bm{v}_{i,j}^{t} \vert i\in \mathcal{M}_t,j\in \mathcal{N}_i\}$).
        \FOR{each client $i$ in $\mathcal{M}_t$}
        \STATE Server sends back updated local task-specific adapters $\{\bm{v}_{i, j}^{t} \vert j\in \mathcal{N}_i\}$.
        \ENDFOR
        \ENDFOR
        \STATE \textbf{return} Global task-specific adapters $\bm{v}_{j}^{T}$ for each task $j \in \{1, \ldots, N\}$.
    \end{algorithmic}
\end{algorithm}

\begin{algorithm}[t]
    \caption{FL-TAC Local Fine-tuning}
    \label{alg:local}
    \textbf{Inputs}:Pre-trained model $\bm{\theta}$; Number of local steps $\tau$; Local task-specific adapters $\{\bm{v}_{i,j}^{t-1} \vert j\in \mathcal{N}_i\}$. \\
    \textbf{Parameters}: Local dataset $D_{i}^{j}$ for $j \in \mathcal{N}_i$; Learning rate $\eta$.
    \begin{algorithmic}[1] 
        \FOR{each task $j \in \mathcal{N}_i$}
        \STATE Initiates $\bm{v}_{i,j, 0}^t = \bm{v}_{i, j}^{t-1}$.
        \FOR{$k=1$ to $\tau$}
        \STATE Updates task-specific adapter: $\bm{v}_{i,j, k}^t = \bm{v}_{i,j, k-1}^t - \eta \nabla{\bm{v}_{i,j, k-1}^t} \mathcal{\hat{F}}_{i,j}(\bm{v}_{i,j, k-1}^t, D_{i}^{j})$.
        \ENDFOR
        \STATE Updates task-specific adapter: $\bm{v}_{i,j}^{t}=\bm{v}_{i,j, \tau}^t$.
        \ENDFOR
        \STATE \textbf{return} Updated local task-specific adapters $\bm{v}_{i,j}^{t}$ for each task $j \in \mathcal{N}_i$.
    \end{algorithmic}
\end{algorithm}

\begin{algorithm}[h]
    \caption{FL-TAC Global Clustering and Aggregation}
    \label{alg:global}
    \textbf{Inputs}: The received clients' task-specific adapters $\{\bm{v}_{i,j}^{t} \vert i\in \mathcal{M}_t,j\in \mathcal{N}_i\}$. \\
    \textbf{Parameters}: Number of downstream tasks $N$. 
    \begin{algorithmic}[1] 
        \STATE Initializes $\mathcal{V}=\{\bm{v}_{i, j}^{t} \vert i\in \mathcal{M}_t,j\in \mathcal{N}_i\}$
        \STATE Performs $K$-means algorithm to split $\mathcal{V}$ into $N$ clusters $\mathcal{V}_1,\mathcal{V}_2,\dots,\mathcal{V}_N$.
        \FOR{$n=1$ to $N$}
        \FORALL{$\bm{v}_{i, j}^{t} \in \mathcal{V}_n$}
        \STATE $\bm{v}_{j}^{t}=\sum_{\bm{v}_{i, j}^{t} \in \mathcal{V}_n} \bm{v}_{i, j}^{t} / \lvert \mathcal{V}_n \rvert$.
        \STATE $\bm{v}_{i, j}^{t} = \bm{v}_{j}^{t}$.
        \vspace{5pt}
        \ENDFOR
        \ENDFOR
       \STATE \textbf{return}  Global task-specific adapters $\bm{v}_{j}^{t}$ for each task $j \in \{1, \ldots, N\}$,
        \STATE \hspace{\algorithmicindent} \hspace{0.52cm} Aggregated local task-specific adapters $\{\bm{v}_{i,j}^{t} \vert j\in \mathcal{N}_i\}$ for each client $i\in \mathcal{M}_t$.
    \end{algorithmic}
\end{algorithm}

The FL-TAC algorithm (Algorithm~\ref{alg:algorithm}) begins with the server initializing a global adapter \(\bm{v}^{0}\) and broadcasting it to all clients in the set \(\mathcal{M}\). Each client \(i\) initializes local task-specific adapters \(\bm{v}_{i,j}^{0}\) for their tasks \(j \in \mathcal{N}_i\). In each communication round \(t\), a subset of clients \(\mathcal{M}_t\) is selected to perform local fine-tuning as described in Algorithm~\ref{alg:local}. More specifically, each selected client \(i\) executes \(\tau\) local training steps with the pre-trained model \(\bm{\theta}\) and local task-specific adapters \(\bm{v}_{i,j}^{t-1}\) for \(j \in \mathcal{N}_i\). Each adapter \(\bm{v}_{i,j}^{t-1}\) is updated iteratively using the local dataset \(D_{i}^{j}\) and a learning rate \(\eta\), resulting in updated local task-specific adapters \(\bm{v}_{i,j}^{t}\) for each local task \(j\).

After the local fine-tuning, each client \(i \in \mathcal{M}_t\) sends its local adapters to the server, which performs global clustering and aggregation as presented in Algorithm~\ref{alg:global}. This process involves the server initializing a collection \(\mathcal{V}\) of the received adapters \(\bm{v}_{i,j}^{t}\) for each client \(i\in \mathcal{M}_t\) and their local tasks \(j \in \mathcal{N}_i\). The \(K\)-means algorithm is then applied to cluster \(\mathcal{V}\) into \(N\) clusters \(\mathcal{V}_1, \mathcal{V}_2, \ldots, \mathcal{V}_N\). Adapters within the same cluster \(n\) are aggregated to form \(\bm{v}_{n}^{t}\) as a new global task-specific adapter for the \(n\)-th task involved in the system, and each local adapter in the cluster is updated to this new aggregated global adapter. Therefore, at the end of the \(t\)-th communication round, the server obtains the global task-specific adapters \(\bm{v}_{n}^{t}\) for each task \(n\) and returns the updated local task-specific adapters corresponding to local tasks \(j \in \mathcal{N}_i\) to each client \(i \in \mathcal{M}_t\). This process is iterative over \(T\) communication rounds, ultimately yielding global task-specific adapters \(\bm{v}_{n}^{T}\) as approximated optimal \(\bm{v}_{n}^{*}\) for each of the \(N\) tasks within the system. Figure~\ref{aggregation_complete} shows the process of FL-TAC algorithm with three example clients, each of which tackles two local tasks.

\vspace{-5pt}
\section{Experiment}
In this section, experiment settings and results for evaluating the performance of the proposed FL-TAC algorithm are presented in details.

\subsection{Experiment Setup}

\textbf{Datasets.} The evaluation is conducted across various datasets, including Databricks-Dolly-15k for text generation, several GLUE datasets for text classification, and CIFAR-10 and CIFAR-100 for image classification~\citep{wang2019glue}. Databricks-Dolly-15k encompasses eight tasks, including Brainstorming, Summarization, Classification, Creative Writing, Information Extraction, Closed Question Answering (QA), Open QA, and General QA. The GLUE datasets considered in this study are Stanford Sentiment Treebank Binary Classification (SST2), Microsoft Research Paraphrase Corpus (MRPC), Quora Question Pairs (QQP), Question-answering Natural Language Inference (QNLI), and Recognizing Textual Entailment (RTE).

\textbf{Base Models.} The tasks of text generation, text classification, and image classification are handled by distinct pretrained models. Specifically, the Llama-7B model~\citep{touvron2023llama} is utilized for text generation, the Bert model~\citep{devlin2018bert} for text classification, and the Vision Transformer (ViT) pretrained on the ImageNet-21k dataset~\citep{dosovitskiy2020image} is employed for image classification.

\textbf{Data Distribution.} For simulating the real-world scenario where each client possesses unique data distribution, Dirichlet Partition, following the work of \citet{hsu2020federated}, is taken into consideration. For each downstream task, the training data is allocated to clients according to the Dirichlet distribution $Dir(\alpha)$. Note that if a client's value from the Dirichlet distribution is below the 0.01 threshold, the task's data will not be allocated to that client. The total number of local clients is 10 and $\alpha$ is set to 0.5 for all downstream tasks. Detailed data distribution can be found in Figure~\ref{data distribution}.

\textbf{Baselines.} For fair and comprehensive evaluation, a range of standard baselines from centralized learning (CL) and FL are selected for comparison. Specifically, \textbf{Llama-7B} and \textbf{Llama-7B-LoRA} are CL models, with Llama-7B representing a well-recognized open-source LLM and Llama-7B-LoRA is its variant fine-tuned with LoRA rank 1 on the same dataset as our method. In terms of the FL method, the baseline \textbf{FedIT}~\citep{zhang2023towards} is considered, which involves fine-tuning Llama-7B with LoRA rank 16 through the FL system.

\begin{table}[htbp]
\caption{Evaluation on Databricks-Dolly-15k dataset}
\begin{center}
\resizebox{\textwidth}{!}{
\begin{tabular}{@{}M{1.5cm} c|M{1.5cm}M{1.5cm}M{1.5cm}M{1.5cm}M{1.5cm}M{1.5cm}M{1.5cm}M{1.5cm}|c|M{2.1cm}@{}}
\toprule[1.5pt]
Training Setting& Method &  Brain\hyp{}storming & Classi\hyp{}fication & Closed QA & Creative Writing & General QA & Information Extraction & Open QA & Summari\hyp{}zation &Average &  Trainable Parameters\\
\midrule
\multirow{2}{*}{\parbox{1.5cm}{\centering CL}} 
& LLama-7B      &0.830&0.640&0.725&0.860&0.930&0.550&0.963&0.661 & 0.770 & 6,607,884,288\\
& Llama-7B-LoRA         & 0.890&0.799&0.978&0.930&0.910&0.880&0.970&0.880& 0.905 & 4,259,840\\
\midrule 
\multirow{2}{*}{\parbox{1.5cm}{\centering FL}}
& FedIT & 0.340 & 0.691 & 0.768 & 0.590 & 0.740 & 0.880 & 0.770 & 0.592 & 0.671 & 8,519,680 \\
& FL-TAC          & \textbf{0.600}&\textbf{0.731}&\textbf{0.855}&\textbf{0.700}&\textbf{0.710}&\textbf{0.770}&\textbf{0.610}&\textbf{0.661}&\textbf {0.705}& \textbf{4,259,840}\\
\bottomrule[1.5pt]
\end{tabular}
}
\label{Databricks-Dolly-15k}
\end{center}
\end{table}

\subsection{Experiment Results}

This subsection presents and analyzes the experiment results obtained under the above experiment settings.

\subsubsection{Performance Analysis for Downstream Tasks}

Table~\ref{Databricks-Dolly-15k} demonstrates the performance of various methods on the Databricks-Dolly-15k dataset across multiple tasks. The evaluation follows previous works by having the GPT-4 model score the answers provided by different methods~\citep{zhang2023towards}. The proposed FL-TAC algorithm surpasses the FedIT baseline in most tasks, except for information extraction, general QA, and open QA. In the information extraction task, it is noteworthy that the performance of the Llama-7B model, centralized fine-tuned with LoRA, is comparatively lower than in other tasks. This observation suggests a potential need for incorporating task-specific adapters and knowledge exchange across tasks for certain cases. For the other two QA tasks, such performance could be attributed to the exceptional language generation ability of the base Llama model, which exhibits good responses to these two types of tasks even without fine-tuning. Figure~\ref{radar} provides a clear visualization of the performance of these methods on the Databricks-Dolly-15k dataset, presenting a radar chart that highlights their respective capabilities.

\begin{table}[htbp]
\caption{Evaluation on GLUE \& Image datasets}
\begin{center}
\resizebox{0.8\textwidth}{!}{
\begin{tabular}{@{}M{1.5cm} c|ccccc|cc|M{2.5cm}|M{2.5cm}@{}}
\toprule[1.5pt]
\multirow{2}[1]{*}{\parbox{1.5cm}{\centering Training\\Setting}}& \multirow{2}[2]{*}{Method} & \multicolumn{5}{c|}{\uline{\hspace{2.5cm} \textbf{GLUE} \hspace{2.5cm}}} &  \multicolumn{2}{c|}{\uline{\hspace{1cm} \textbf{Image} \hspace{1cm}}} & \multirow{2}[1]{*}{\parbox{2.5cm}{\centering Trainable\\Parameters (Bert)}}&\multirow{2}[1]{*}{\parbox{2.5cm}{\centering Trainable\\Parameters (ViT)}}\\
&   &QNLI & QQP &SST-2 & RTE & MRPC & CIFAR10 & CIFAR100 & \\
\midrule
\multirow{4}{*}{\parbox{1.5cm}{\centering CL}} 
& Bert      & 0.901 & 0.875 &0.923 & 0.657 & 0.840&  - & - & 108,311,810&  -\\
& Bert-LoRA         & 0.792 & 0.813 & 0.884& 0.552 & 0.685 & - & - & 192,010 &- \\
& ViT      & - & - & - & - & - &  0.987 & 0.920 & - & 86,862,568\\
& ViT-LoRA  & - & - & - & - & - & 0.950& 0.899 & - & 36,864 \\
\midrule 
\multirow{2}{*}{\parbox{1.5cm}{\centering FL}}
& FedIT & 0.670& 0.510 &0.652 &0.524 & 0.520& 0.935& 0.854& 614,432 & 294,912 \\
& FL-TAC          & \textbf{0.787} & \textbf{0.807} & \textbf{0.874} & \textbf{0.532} & \textbf{0.673}& \textbf{0.946}& \textbf{0.884}& \textbf{192,010}&\textbf{36,864}\\
\bottomrule[1.5pt]
\end{tabular}
}
\label{GLUE&CIFAR}
\end{center}
\end{table}

Furthermore, Table~\ref{GLUE&CIFAR} displays the accuracy achieved by the aforementioned methods on the GLUE dataset and image datasets. The proposed FL-TAC algorithm also demonstrates superior performance against the FedIT baseline in terms of both accuracy and communication efficiency.

\subsubsection{Performance of Clustering at the Central Server}

\begin{figure}[h]
\begin{center}
\includegraphics[width=\textwidth]{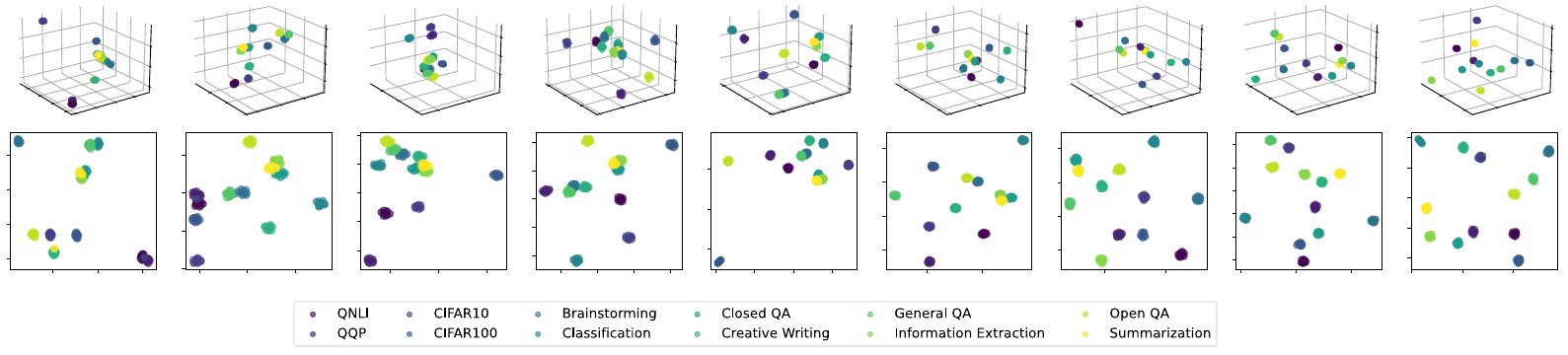}
\end{center}
\caption{Visualization of clustering results from epoch 1 to epoch 9.}
\label{umap}
\end{figure}

The visualization of clustering results using UMAP~\citep{mcinnes2018umap} dimension reduction after the $K$-means clustering step is presented in Figure~\ref{umap}. As the training epochs advance, the categorization becomes more distinct and the distances between each cluster increase. This progression indicates the successful clustering of task-specific adapters through the training process of FL-TAC.


\subsubsection{Analysis of Communication Cost and Local Training Cost}

The trainable parameters, as shown in Table~\ref{Databricks-Dolly-15k} and Table~\ref{GLUE&CIFAR}, display a comparison of the quantity of training parameters required by each method. Notably, the LoRA technique drastically decreases the number of training parameters required for fine-tuning. Also, when compared with FedIT, the training parameter transmitted by FL-TAC is lower, which is a result of our strategic approach of training task-specific adapters with a lower LoRA rank.

\vspace{-5pt}
\section{Conclusion}

In conclusion, this work enhances fine-tuning large pre-trained models within FL by reducing communication cost and improving task adaptation performance with the presence of various downstream tasks. The introduced FL-TAC method employs low-rank task-specific adapters for each client and performs clustering and aggregation of similar adapters on the server. The extensive experiments of fine-tuning via the proposed FL-TAC method demonstrate improved communication efficiency and enhanced task adaptation performance compared to the single-adapter baseline, contributing to a unified, scalable FL framework for pre-trained model fine-tuning with both efficient communication and effective multi-task generalization. Future research directions involve investigating unresolved issues, such as the correlation between the required LoRA rank and the capabilities of base models.



\vspace{-5pt}
\section{Acknowledgement}
This work was supported by Shenzhen Key Laboratory of Ubiquitous Data Enabling (ZDSYS20220527171406015), the National Key R\&D Program of China under Grant No.2022ZD0160504 and Meituan.

\bibliography{iclr2024_conference}
\bibliographystyle{iclr2024_conference}

\end{document}